\documentclass[acmlarge, screen]{acmart}

\renewcommand\footnotetextcopyrightpermission[1]{}
\begin{document}

\title{Toward Continuous Neurocognitive Monitoring: Integrating Speech AI with Relational Graph Transformers for Rare Neurological Diseases}

\author{Raquel Norel}
\affiliation{
  \institution{IBM Research}
  \city{Yorktown Heights}
  \state{NY}
  \country{USA}
}
\email{[RNorel@gmail.com]}

\author{Michele Merler}
\affiliation{
  \institution{IBM Research}
  \city{Yorktown Heights}
  \state{NY}
  \country{USA}
}
\email{[mimerler@us.ibm.com]}

\author{Pavitra Modi}
\affiliation{
  \institution{IBM Software}
  \city{[Markham]}
  \country{Canada}
}
\affiliation{
  \institution{University of New Brunswick}
  \city{Fredericton}
  \state{NB}
  \country{Canada}
}
\email{[Pavitra.Modi@ibm.com/pmodi@unb.ca]}

\begin{abstract}
Patients with rare neurological diseases report cognitive symptoms—``brain fog''—invisible to traditional tests. We propose continuous neurocognitive monitoring via smartphone speech analysis integrated with Relational Graph Transformer (RELGT) architectures. Proof-of-concept in phenylketonuria (PKU) shows speech-derived ``Proficiency in Verbal Discourse'' correlates with blood phenylalanine ($\rho=-0.50$, $p<0.005$) but not standard cognitive tests (all $|r|<0.35$). RELGT could overcome information bottlenecks in heterogeneous medical data (speech, labs, assessments), enabling predictive alerts weeks before decompensation. Key challenges: multi-disease validation, clinical workflow integration, equitable multilingual deployment. Success would transform episodic neurology into continuous personalized monitoring for millions globally.
\end{abstract}


\maketitle

\section{The Monitoring Gap}

Patients with rare neurological diseases consistently report cognitive difficulties undetectable by clinical assessments~\cite{Bilder2016}. In phenylketonuria (PKU), adults describe ``brain fog'' and working memory deficits~\cite{McWhirter2023} yet score normally on neuropsychological tests~\cite{Romani2022}. Metabolic decompensation progresses undetected for weeks between quarterly blood tests.

A key challenge in PKU research is how to bridge subjective patient experience with objective, actionable biomarkers. Traditional assessment methods fail short in several ways: episodic clinic visits miss the everyday fluctuations that patient reports, standard neurocognitive tests lack ecological validity and do not capture how PKU affects real-world functioning, and the data that is collected often remains siloed across different formats and sources. This makes it difficult for clinicians to understand how cognitive load, mood, fatigue, and metabolic changes interact over time, and prevents the development of continuous, meaningful indicators of a patient's true neurological state.

\begin{figure}[H]
\centering
\includegraphics[width=1\linewidth]{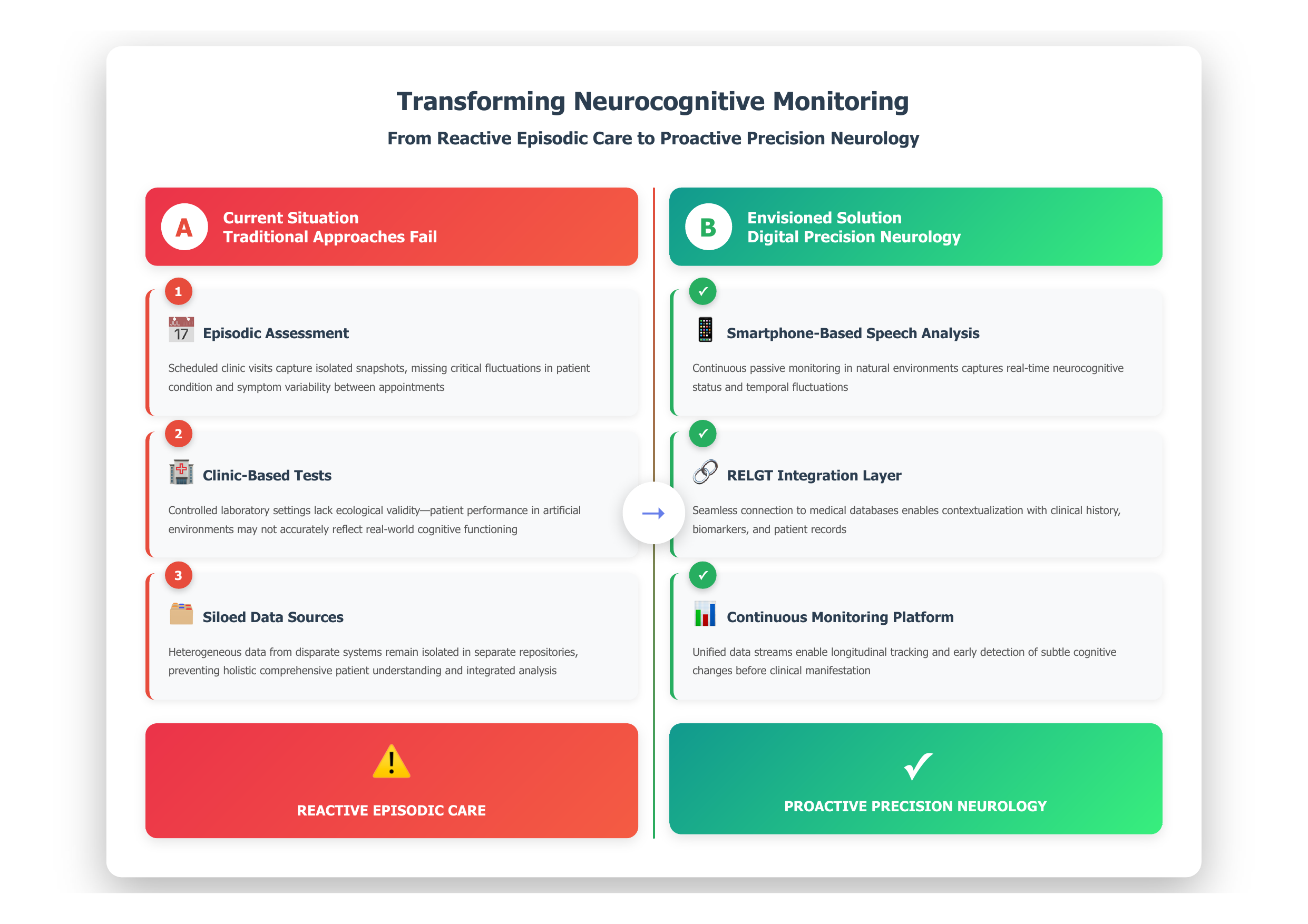}
\caption{Contrasting Traditional and Digital Approaches to Neurocognitive Monitoring. Panel A illustrates limitations of traditional episodic assessment, clinic-based testing, and siloed data sources that result in reactive care. Panel B presents an envisioned solution using smartphone-based speech analysis integrated with medical databases via RELGT, enabling continuous monitoring and proactive precision neurology through real-time data capture in remote, natural environments.} 
\label{fig:schema}
\end{figure}

 Natural language processing quantifies spontaneous speech as a neurocognitive biomarker~\cite{Eyigoz2020,Norel2020}; Relational Graph Transformers (RELGT) can integrate heterogeneous medical data at scale~\cite{Dwivedi2025}. \textbf{We envision smartphone-based speech analysis integrated with medical databases via RELGT, enabling continuous neurocognitive monitoring—transforming reactive episodic care into proactive precision neurology. Figure \ref{fig:schema} illustrates the envisioned schematic shift.}

\section{Vision: Integrated Framework}

\textbf{Component 1: Speech Biomarkers.} Spontaneous speech engages executive control, semantic retrieval, working memory, and pragmatic language—vulnerable in frontal-subcortical dysfunction~\cite{Alexander2006}. AI analysis of 60-second narratives yields features capturing discourse complexity, coherence, and syntax. Smartphone-based capture occurs continuously in natural environments.

\textbf{Component 2: RELGT Architecture.} Medical data is mostly stored and represented as relational entities: patients$\rightarrow$tests$\rightarrow$treatments$\rightarrow$symptoms. RELGT~\cite{Dwivedi2025} can handle heterogeneous graphs where nodes encode data types (speech, labs, medications) and edges capture relationships (temporal, causal). Unlike GNNs that limited by information bottlenecks, RELGT employs hybrid attention, enabling direct integration across multiple hops.

\textbf{Component 3: Predictive Alerts.} Learning nonlinear cross-modal relationships identifies baseline deviations and generates risk-stratified alerts. Example: Week 3 speech decline triggers an assessment revealing metabolic elevation before scheduled Week 8 blood test — intervention 4+ weeks earlier than with traditional approaches.

\textbf{Proof-of-Concept:} PKU analysis ($n=42$ vs 41 controls)~\cite{Waisbren2025} identified 23 linguistic features aggregated into ``Proficiency in Verbal Discourse'' that:

\begin{itemize}
\item Correlated with phenylalanine ($\rho=-0.50$, $p<0.005$) and tyrosine ($\rho=0.44$)
\item Showed zero correlation with WAIS-IV (all $r<0.17$, $p>0.1$)
\item Captured complexity, detail, coherence, context, emotion (ChatGPT-4 analysis)
\end{itemize}

Despite a 40\% scoring from speech biomarkers that was clinically significant for working memory deficits and a 45\% reporting for neurocognitive burden, standard tests administered to the same individuals resulted in ``normal'' outcomes. Continuous speech analysis has therefore capability to bridge lived experience and assessment measurements.

\section{Key Research Challenges}

\subsection{Challenge 1: Multi-Disease Validation}

A major challenge lies in ensuring that speech-based models generalize across disorders with fundamentally different pathophysiology. Parkinson's disease involves hypophonia and speech fluctuations tied to medication timing. Huntington's disease reflects CAG-repeat-driven degrneration and progressive motor-cognitive decline. Wilson's disease presents with dysarthria linked to copper accumulation. Each condition affects speech through different mechanisms, yet all share speech as an accessible, real-time window into brain function. Research must explore cross-disease learning to determine which speech patterns are disease-specific, which are shared across conditions, and how to build models that remain robust despite this biological diversity. Success can be achieved by the developemnt of biomarkers with hih correlation levels across deseases, in ened-to-end speech based models with high precision across conditions, or ideally both. 


\subsection{Challenge 2: Scalable Integration}

Another key challenge consists is integrating millions of heterogenous speech and clinical records while accounting for temporal evolution overal extended time periods (potentially decades) and missing/incomplete data. Large-scale modeling requires architectures that can handle millions of nodes, irregular sampling, and long-term patient trajectories withoug sacrificing interpretability. Research directions include developing hybrid attention mechanisms optimized for massive graphs, learning individualized temporal trajectories, applying federated learning to protect patient privacy, and designing explainable attention weights that reveal which signals drive predictions. Scalable, transparent systems will be essential for deploying speech-based monitoring across real-world clinical populations. 


\subsection{Challenge 3: Clinical Workflow}

Digital biomarkers only succeed when they fit naturally into clinical routines and do not add burden for patients or clinicians. Many promising technologies fail at this step, not becayse they lack accuracy, but because they disrupt workflows or generate unmanageable alert volumes. Research must focus on seamless integration with clinical systems through standards like FHIR, the design of risk-adaptive sampling strategies that minimize patient effort, careful calibration of alert thresholds to avoid overload, and studies that evaluate clinician trust and usability. Ensuring low-friction adoption is essential for turning speech-based biomarkers into real clinical tools. 


\subsection{Challenge 4: Health Equity}

The final challenge is ensuring that speech-based biomarkers do not widen existing health disparities. Models trained on narrow populations risk failing for speakers of different languages, dialects, or with limited access to technology. Research must therefore prioritize multilingual datasets spanning several languages and domain adaptation \cite{Guan2022}/transfer learning \cite{zhuang_transfer_survey2020} techniques for underrepresented languages, develop alternative interfaces for users with speech or motor limitations, support on-device processing for those with limited connectivity, and conduct continuous bias auditing to identify and correct systematic errors. Equity-focused design is essential for ensuring that these tools benefit all patients, not just those represented in early datasets. 


\section{Towards Precision Neurology}

Solving the above challenges would shift neurological care from episodic snapshots to continuous monitoring, from population norms to individualized trajectories, and from reactive responses to predictive insight. Over 50 million individuals are affected by neurodegenerative diseases~\cite{Feigin2019}; hundreds of thousands have neurometabolic conditions~\cite{NguengangWakap2020}, yet many of their symptoms remain invisible to current monitoring tools. 

Speech provides a natural window into brain functions through non-intrusive devices that patients already naturally interact with every day. With the convergence of advanced models such as relational graph transformers, widespread smartphone access, and increasingly rich medical databases, this vision could become achievable.  \textbf{We call on the community to tackle those challenges—transforming how we monitor, predict, and preserve brain health for millions globally.}


\bibliographystyle{ACM-Reference-Format}
\bibliography{acmart}

\end{document}